\pdfoutput=1

\documentclass[11pt]{article}

\usepackage[preprint]{acl}

\usepackage{times}
\usepackage{latexsym}
\usepackage{bm}
\usepackage{array}
\usepackage{times}
\usepackage{makecell}
\usepackage{latexsym}
\usepackage{graphicx} 
\usepackage{algorithm}
\usepackage{algpseudocode}
\usepackage{amsmath} 
\usepackage[utf8]{inputenc}
\usepackage{graphicx}
\usepackage{booktabs}
\usepackage{tabularx}
\usepackage{amssymb}
\usepackage{xspace}

\usepackage{amsmath,amsfonts,bm}

\def\eqref#1{equation~\ref{#1}}

\def\1{\bm{1}}

\def\vx{{\bm{x}}}

\DeclareMathAlphabet{\mathsfit}{\encodingdefault}{\sfdefault}{m}{sl}
\SetMathAlphabet{\mathsfit}{bold}{\encodingdefault}{\sfdefault}{bx}{n}

\def\gL{{\mathcal{L}}}

\def\gV{{\mathcal{V}}}

\usepackage[normalem]{ulem}
\usepackage{multirow}
\usepackage{colortbl}     

\usepackage[T1]{fontenc}

\usepackage[utf8]{inputenc}

\usepackage{microtype}

\usepackage{inconsolata}

\usepackage{amsmath}
\usepackage{amsfonts}
\usepackage{multirow}

\newcommand{\ie}{\emph{i.e.},\xspace}

\newcommand{\fakeparagraph}[1]{\vspace{1mm}\noindent\textbf{#1}}

\usepackage{graphicx}

\title{Get Confused Cautiously: Textual Sequence Memorization Erasure with Selective Entropy Maximization}

\author{Zhaohan Zhang\thanks{Preprint. Zhaohan Zhang is the corresponding author.}, Ziquan Liu, Ioannis Patras \\
        Queen Mary University of London, London, UK \\ 
        \texttt{\{zhaohan.zhang, ziquan.liu, i.patras\}@qmul.ac.uk} 
        }

\begin{document}
\maketitle
\begin{abstract}
Large Language Models (LLMs) have been found to memorize and recite some of the textual sequences from their training set verbatim, raising broad concerns about privacy and copyright issues when using LLMs.
This Textual Sequence Memorization (TSM) phenomenon leads to a high demand to regulate LLM output to prevent it from generating certain memorized text to meet user requirements.
However, our empirical study reveals that existing methods for TSM erasure fail to forget massive memorized samples without substantially jeopardizing the model utility.
To achieve a better trade-off between the effectiveness of TSM erasure and model utility in LLMs, our paper proposes a new framework based on Entropy Maximization with Selective Optimization (EMSO), where the updated weights are chosen with a novel contrastive gradient metric without any participation of additional model or data.
Our analysis shows that training with the entropy maximization loss has a more stable optimization process and better keeps model utility than existing methods.
The contrastive gradient metric localizes the most influential weight for TSM erasure by taking both the gradient magnitude and direction into consideration.
Extensive experiments across three model scales demonstrate that our method excels in handling large-scale forgetting requests while preserving model ability in language generation and reasoning.

\end{abstract}

\section{Introduction}
Large Language Models (LLMs) are a series of transformers-based models pre-trained on an enormous corpus with trillions of tokens, achieving human-level performance on language abilities. \cite{vaswani2017attention, brown2020language, touvron2023llama, achiam2023gpt}.
While the utility of LLMs greatly benefits from scaling laws \cite{kaplan2020scaling}, recent studies reveal that LLMs have \textit{Textual Sequence Memorization (TSM)}, i.e., memorizing and emitting training samples verbatim, including Personally Identifiable Information (PII) and copyrighted content \cite{carlini2021extracting, huang2022large, jagielski2022measuring}.
This phenomenon raises serious concerns about violating the regulation of the right to be forgotten (RTBF) \cite{mantelero2013eu, graves2021amnesiac}.
Hence, erasing TSM from LLMs is in great demand to protect PII and intellectual property.

\begin{figure}[t]
  \centering
  \includegraphics[width=0.95\linewidth]{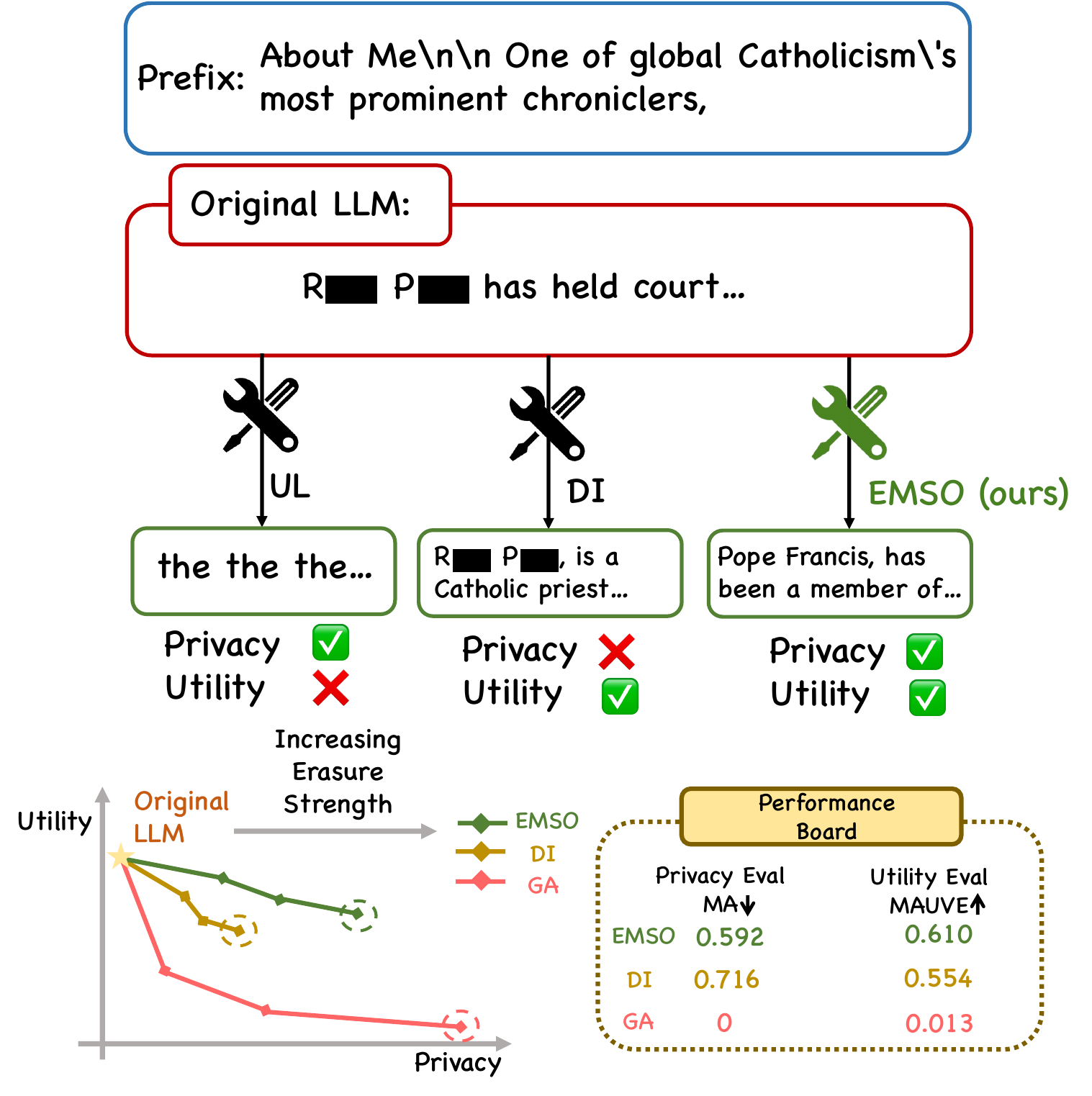}
  \vspace{-0.35cm}
  \caption{Illustration of erasure-utility trade-off with the example of three methods: Gradient Ascent (GA) \cite{jang2023knowledge}, Deliberate Imagaination (DI) \cite{dong2024unmemorization} and our EMSO. The top figure shows the exemplary erasure-utility trade-off with different cases. The bottom figure demonstrates the quantitative erasure-utility trade-off with the correlation between TSM metric (Memorization Accuracy, MA) and generation quality metric (MAUVE). }
  \label{fig:intro} 
  \vspace{-15pt}
\end{figure}

There are two major types of memorization erasure for LLMs in existing literature.
1) \emph{Knowledge erasure} focuses on the removal or modification of abstract knowledge, such as factual associations \cite{wang2024large, mengmass} or hazardous knowledge \cite{liwmdp, liu2024towards} within LLMs.
These works evaluate the model acquisition of unwanted knowledge using \textit{question-answering} tasks in the form of classification.
For example, Weapons of Mass Destruction Proxy (WMDP) benchmark \cite{liwmdp} constructs a dataset of \textit{multiple-choice questions} to serve as a proxy measurement of hazardous knowledge and evaluate 
the efficacy of knowledge erasure with classification accuracy drop. 2) \emph{Textual sequence memorization erasure} prevents the model from generating sequences with high verbatim similarity with training data \cite{carlini2021extracting, carlini2022quantifying, barbulescu2024each}.
Compared with knowledge memorization in classification tasks, TSM has a closer relationship with current privacy and copyright challenges in LLMs given the fact that the most popular LLMs are generative models. 
Thus, recent works within the scope of TSM erasure are commonly evaluated on \textit{open-end generation} tasks (\ie continuation based on given prefix) \cite{jang2023knowledge, kassem2023preserving, yao2023large}.

This work focuses on erasing TSM of user-designated data from LLMs. 
Memorized data is deeply tied to general language modelling \cite{huang2024demystifying}, making it hard to remove without reducing model utility. 
As shown in Fig.\ref{fig:intro}, current methods often either erase TSM or maintain model utility, creating an erasure-utility trade-off dilemma.
Existing erasure methods rely on references like memorized models \cite{ilharco2022editing, li2023contrastive, eldan2023s} or retained data \cite{liu2022continual, wang2023kga} to manage model utility.
However, these references introduce challenges: memorized models can compromise privacy, while acquiring and maintaining retained data can be impractical.
Furthermore, \citet{maini2024tofu} has observed instances of model collapse when attempting to erase extensive memorized data in one operation.

Recognizing the limitations of previous work mentioned above, we aim to improve the TSM erasure while preserving model utility with three desired properties:
(\textit{i}) \textbf{erasing without} involving a memorized model to avoid privacy issues;
(\textit{ii}) \textbf{erasing with} only access to forget set without a retain set;
(\textit{iii}) \textbf{erasing with} a large-scale forget set to accommodate large-scale erasure requests.
To tackle these challenges, we design a novel framework for TSM erasure, \emph{entropy maximization with selective optimization} (EMSO). The proposed objective function is to increase the entropy of the predictive distribution on a forget set to encourage more diverse output instead of penalizing the generation of memorized tokens.
Moreover, to keep the original model utility, we apply a \emph{minimally invasive surgery} to the model by only updating the most significant weights for entropy maximization.
To be specific, we design a novel reference-free metric that takes both gradient magnitude and direction into consideration.
This metric helps to locate weights that contribute positively to entropy maximization while negatively to token memorization.
Extensive experiments show that our method achieves a better erasure-utility trade-off when processing massive erasure requests compared with recent baselines.
Our contribution is summarized as follows:
\begin{itemize}
    \item We introduce a reference-free optimization objective to enhance forgetting of large-scale memorized data in LLMs. This objective aims to increase predictive distribution entropy, which proves to be a more stable optimization target compared to commonly used gradient ascent and label smoothing methods, supported by both theoretical analysis and empirical findings.

    \item We propose a selective optimization approach that only updates salient weights selected by a contrastive gradient metric to achieve a better erasure-utility trade-off. The metric prefers a weight that is significant only for entropy maximization but not memorization based on gradient magnitude and direction.
    
    \item Our empirical study demonstrates that our EMSO method for TSM erasing achieves the best trade-off between information leakage and model utility on a large-scale forget dataset across various metrics and model sizes. 
\end{itemize}
\vspace{-0.3cm}



\begin{figure*}[t]
\centering

\includegraphics[width=0.95\textwidth]{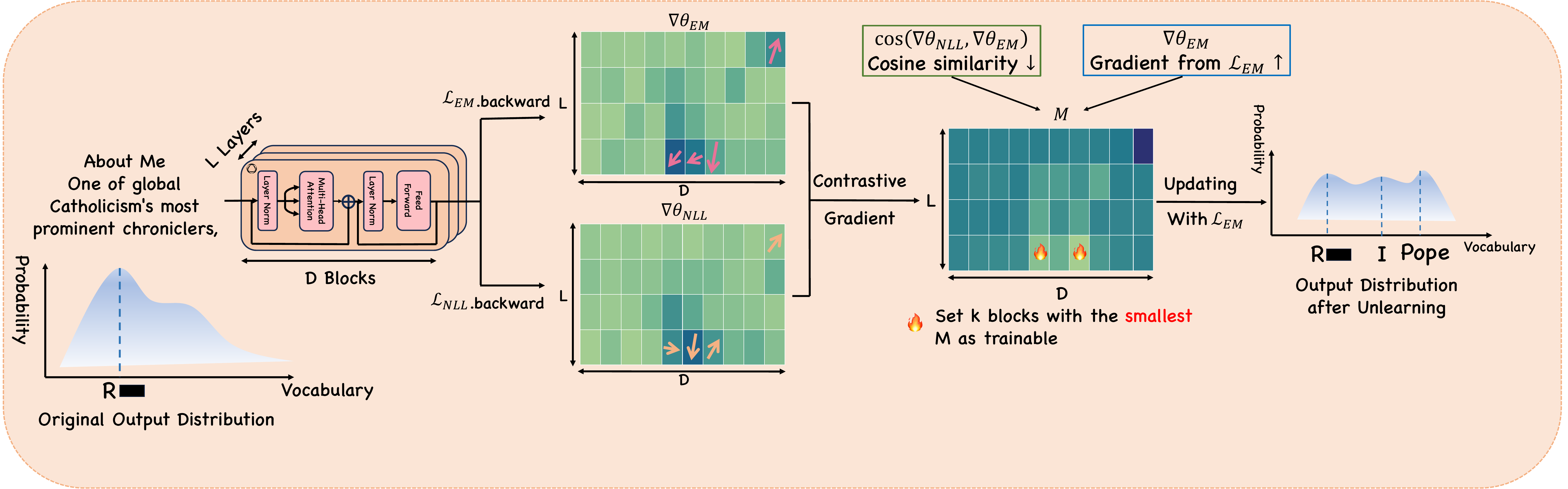}
\vspace{-0.3cm}
\caption{Framework of EMSO. We select the most significant weights for forgetting based on the gradient magnitude and direction so that the weight is effective at entropy maximization instead of data memorization.  }
\label{fig:framework}
\vspace{-0.5cm}
\end{figure*}

\section{Related Works}


\fakeparagraph{Knowledge Unlearning for LLMs.}
Machine unlearning aims at removing model memorization about sensitive data.
In contrast to traditional unlearning approaches in classification tasks \cite{bourtoule2021machine, chundawat2023can, jia2023model}, the concept of machine unlearning in generative LLMs shifts focus to the characteristics of model output.
Specifically, it focuses on 
mitigating harmful or biased information in generated content.
Abstract harmful knowledge is one of the targets for LLM unlearning, which bears similarities with safety alignment but primarily uses negative samples \cite{liwmdp, liu2024towards, yao2023large}.
Question-answering-based benchmarks such as TOFU \cite{maini2024tofu} and WMDP \cite{liwmdp} are established for evaluating model acquisition of the knowledge.
Given the objective and testbase, rejection-based methods such as Direct Preference Optimization (DPO) \cite{rafailov2024direct} are suitable for encouraging the model to answer malicious questions.
The evaluation metrics for quantifying hazardous knowledge in LLMs include accuracy on malicious multiple-choice questions \cite{liwmdp} and GPT-as-a-Judge score \cite{liu2024towards}.

\fakeparagraph{TSM Erasure in LLMs}
TSM refers to LLMs' ability to memorize and emit training samples verbatim, which is an undesired attribute to be erased/unlearned \cite{carlini2021extracting, carlini2022quantifying}.
Different from undesired knowledge, TSM is defined over certain training data points.
For evaluation, the updated model is asked to generate continuation based on the prefix of memorized data.
Training Data Extraction Challenge \footnote{https://github.com/google-research/lm-extraction-benchmark} serves as a persuasive benchmark for probing TSM in GPT-Neo model family. 
Concurrent work MUSE \cite{shi2024muse} provides news and books corpus for evaluating verbatim memorization. 
The extent of memorization is calculated by the similarity between the output (before or after decoding) of the original model and updated model \cite{jang2023knowledge, barbulescu2024each, wang2024selective}.
Gradient ascent \cite{jang2023knowledge} is a straightforward approach for erasing TSM by maximizing the probability of wrong prediction for samples in the forget set.
Other objectives such as Deliberate Imagination \cite{dong2024unmemorization} and Negative Preference Optimization \cite{zhang2024negative} are proposed to avoid model collapse during model updation.
Recent works also try to localize the specific model units where the TSM is stored.
For example, \citet{wu2023depn} localizes privacy neurons with gradient integration and deactivates the identified neurons to protect private information.
\citet{jia2023model, fan2024salun} indicate weight saliency is informative for locating model units that are beneficial to unlearning. 
Our work, as a localization-informed method for TSM erasure, differs from the above-mentioned works in (\textbf{\textit{i}}) proposing a new stabler objective for TSM erasure; (\textbf{\textit{ii}}) taking gradient direction into consideration without retain data for localizing important weights.

\section{Methodology}
\vspace{-0.2cm}
EMSO reference-freely removes TSM by selecting weights to be updated by contrastive gradient metric and optimizing towards entropy maximization objective.
The workflow of EMSO is shown in Figure \ref{fig:framework}.

\subsection{Problem Definition}
Let $\vx_i=(x_1,..., x_p, ..., x_{p+q})$ be a textual sequence where $x_{1:p}$ is the prefix and $x_{p+1:q}$ is the original continuation.
Given a forget set $D_f \in D$, where $D$ is the pre-training dataset for an LLM $\theta_o$, the objective of TSM erasure is to obtain an updated model $\theta_u$ which performs exactly the same as a model only trained on $D \textbackslash D_f$, \ie dataset which is obtained by deleting $D_f$ from $D$.
This goal implies that the updated model $\theta_u$ should keep its utility on $D \textbackslash D_f$ as same as the original model $\theta_o$ while showing "unmemorization" effect\footnote{The unmemorization effect refers to model's disability to recite the text sequence in $D_f$ verbatim.} on $D_f$.
Ideally, the updated model $\theta_u$ can be obtained by pre-training an LLM from scratch with $D \textbackslash D_f$.
However, due to the prohibitive computational cost it requires \cite{yao2023large}, such a solution is commonly recognized as unrealistic \cite{liu2024towards, wang2023kga, jang2023knowledge}.
In this work, we aim to directly update $\theta_o$ with access only to forget set $D_f$ to approximate the performance of $\theta_u$ on both $D_f$ and $D \textbackslash D_f$.

\subsection{Entropy Maximization}
\label{subsec:em}

Entropy is the measurement of the uncertainty of a probability distribution $P$.
In the context of LLM generation, a larger entropy on the next token probability $P_\theta(x_i|x_{<i})$ indicates that the model is uncertain about its decision on the current decoding token, leading to a higher probability to select other reasonable tokens and output more diverse content.
Importantly, this diversity helps prevent the model from memorizing specific sequences.
We find entropy maximization is 
Thus, we propose to maximize the entropy of $P_\theta(x_i|x_{<i})$ on $D_f$ by minimizing the following loss function
\setlength{\belowdisplayskip}{1.0pt} \setlength{\belowdisplayshortskip}{1.0pt}
\setlength{\abovedisplayskip}{1.0pt} \setlength{\abovedisplayshortskip}{1.0pt}
\begin{align}
    \mathcal{L}_{EM} &= \frac{1}{q}\sum_{i=1}^{q} \sum_{y \in |\mathcal{V}|}P^i_{\theta,y}\text{log} P^i_{\theta,y} ,
    \\ 
P^i_{\theta,y}&=P(x_{(p+i)}=y|x_{<(p+i-1)};\theta),
\end{align}
where $p,q$ are the lengths of the prefix and continuation, respectively.
$P^i_{\theta,y}$ denotes the probability of predicting the $i$-th token to be $y$, $\mathcal{V}$ is the vocabulary and $|\gV|$ is its cardinality.
Compared with commonly used objectives, we theoretically prove that entropy maximization objective helps stabilize the model updation process during TSM erasure in the following section.

\noindent\textbf{Comparison to Label Smoothing Loss and Gradient Ascent Loss.} 
Label smoothing loss \cite{muller2019does, dong2024unmemorization} and gradient ascent loss \cite{liu2022continual, jang2023knowledge, wang2023kga} have emerged as two popular objectives for TSM erasure.
As a new learning objective, our EM loss is more stable during the optimization. 
The gradient analysis shows that the minimizer of the label smoothing loss is identical to the maximizer of the EM loss. 
For each token $i$, the label smoothing loss is as follows, 
\begin{align}
    \gL_{ls}=-\gamma\sum_{j=1}^{|\gV|}\log \hat p_{ij},
\end{align}
where $\hat p_{ij}=P(x_{(p+i)}=j|x_{<(p+i-1)};\theta)$ for simplicity and $\gamma$ is the hyperparameter of the label smoothing loss. As $\hat p_{ij}$ is the output of the softmax function with $h_{ij}$ as the input, we take the derivative of the loss function with respect to the input $h_{ij}$,

\begin{small}
\begin{align}
    \frac{\partial \gL_{ls}}{\partial h_{ik}}&=-\gamma \sum_{j=1}^{|\gV|}\frac{1}{\hat p_{ij}}\frac{\partial \hat p_{ij}}{\partial h_{ik}}\label{equ:ls_grad}\\
    &=-\gamma\sum_{j\neq k}^{|\gV|}\frac{1}{\hat p_{ij}}(-\hat p_{ij}\hat p_{ik}) - \gamma \frac{1}{\hat p_{ik}}(\hat p_{ik}-\hat p_{ik}^2)\nonumber \\
    &=-\gamma [-(|\gV|-1)\hat p_{ik}+1-\hat p_{ik}]\nonumber \\
    &=-\gamma (1-|\gV|\hat p_{ik}) \nonumber.
\end{align}
\end{small}

It is trivial to get that the minimizer of the function is $\forall k$, $\hat p_{ik}=1/|\gV|$, which is equivalent to the optimum of the maximum entropy loss. 

We next derive the gradient of $\gL_{EM}$ with respect to the logits,
\begin{align}
    \frac{\partial \gL_{EM}}{\partial h_{ik}}=\sum_{j=1}^{|\gV|}(\log \hat p_{ij}+1)\frac{\partial \hat p_{ij}}{\partial h_{ik}}. \label{equ:me_grad}
\end{align}

Comparing the gradient \ref{equ:ls_grad} with the gradient \ref{equ:me_grad}, the only difference is the first term. As $\hat p_{ij}\in[0,1]$, the scale and gradient of $\log \hat p_{ij}$ is much smaller than that of $-1/\hat p_{ij}$, we provide an illustration in Appendix \ref{appd:mini} for reference.
Note that the gradient scale analysis result is also applicable to gradient ascent loss (details are in Appendix \ref{appd:mini}), indicating that the gradient ascent loss also has the risk of unstable optimization. In summary, our entropy maximization loss has the same optimization objective but much more stable gradients compared with the label smoothing loss and gradient ascent.
In Section \ref{subsec:exp}, our experiment results corroborate the gradient analysis. 

\vspace{-0.2cm}
\subsection{Weight Selection with Contrastive Gradient}
To achieve a better trade-off between erasure effectiveness and model utility, we propose to only finetune weights that are salient to forgetting and keep other weights the same as the original to preserve model utility.
We select weights from all attention heads and the multi-layer perceptron (MLP) block in every layer $l$ because they are components of the "residual block" which acts as  communication channels in transformers-based architectures \cite{elhage2021mathematical}.
Inspired by the gradient-based input salient maps \cite{adebayo2018sanity,yona2021revisiting}, we use weight saliency 
$\nabla \theta_{EM} \in \mathbb{R}^{L\times C \times D}$
with respect to $\mathcal{L}_{EM}$ as a metric for selecting influential weights\footnote{L is the number of layers, C is the number of candidate blocks, \ie attention heads and MLP blocks, D denotes the dimension of weight vector. Please note that for simplicity of notation, the denotation assumes that D is the same across layers and components.}:
\begin{equation}
    \nabla \theta_{EM} = \frac{\partial\mathcal{L}_{EM}}{\partial \theta } 
\end{equation}
However, maximizing the entropy of output distribution updates $\theta_o$ towards a more diverse output but not precise "unmemorization". Thus, we design a contrastive gradient strategy to select weights that are both salient with respect to $\mathcal{L}_{EM}$ and contributive to unmemorization.
Taking inspiration from previous works \cite{zhang2023composing, eldan2023s} which train a memorization model on the forget set by minimizing $\mathcal{L}_{NLL} = -\frac{1}{q}\sum_{i=1}^{q}\text{log}(P^i_{\theta,x_{p+i}})$, we consider the gradient direction with respect to $\mathcal{L}_{NLL}$ minimization as "memorization direction".
Thus, we propose an updated metric $M \in \mathbb{R}^{L \times C}$ taking both direction and magnitude into consideration:
\begin{equation}
\begin{split}
    M = \text{cos}(\nabla \theta_{NLL}, &\nabla \theta_{EM})\frac{|\nabla \theta_{EM}|}{\sqrt{D}},\\
    \nabla \theta_{NLL} =&\frac{\partial \mathcal{L}_{NLL} }{\partial \theta }, 
\end{split}
\end{equation}
where cos( $\cdot$ ) is cosine similarity and $|\cdot|$ is $l_1$ norm function.
We scale the $l_1$ norm by $\frac{1}{\sqrt{D}}$ to eliminate the effect of various dimensions of different model components. 
The cosine similarity measures the disagreement of optimization direction between $\mathcal{L}_{EM}$ and $\mathcal{L}_{NLL}$. $|\nabla \theta_{EM}|$ measures the parameter saliency to the optimization of  $\mathcal{L}_{EM}$.
Note that the direction of $\nabla \theta_{NLL}$ represents memorization and the direction of $\nabla \theta_{EM}$ is the updation direction.
If the cosine similarity is large (positive), it means this weight is optimized towards memorization, which is not desirable for a good trade-off. If the cosine similarity is small (negative), this weight is updated towards forgetting.
Therefore, the selected weight should be optimized towards "forget direction" and be salient to $\mathcal{L}_{EM}$, see Figure~\ref{fig:framework} for the illustration.
Thus, we obtain the block-wise weight mask $\textbf{m}$ according to $M$:
\begin{equation}
    \textbf{m} = \textbf{1} (topk(-M)),
\end{equation}
where $\textbf{1}(\text{top}k(\textbf{g}))$ is an element-wise indicator which labels 1 for the top-$k$ element in $\textbf{g}$.
In practice, we empirically observe that setting $k$ to 2 yields sufficiently effective performance. We show the influence of different $k$ in Appendix \ref{appd:k}.
The updating process of the original model $\theta_o$ can be expressed as:
\begin{equation}
    \theta_u \leftarrow \theta_o - \alpha \textbf{m}\odot\nabla\theta_{EM},
\end{equation}
where $\odot$ denotes element-wise product and $\alpha$ is learning rate.
Our experiment selects one batch randomly as input to calculate $\textbf{m}$.

\section{Experiments}

\begin{table*}[ht]
\centering
\renewcommand\arraystretch{1.1}
\resizebox{\textwidth}{!}{
\begin{tabular}{cccccccccccccc}
\toprule \midrule
& \multicolumn{1}{l}{}                                              & \multicolumn{1}{l}{}   & \multicolumn{3}{c}{\textbf{Memorization}} & \multicolumn{5}{c}{\textbf{Language Generation Ability}}       & \multicolumn{3}{c}{\textbf{Ranking}} \\
\cmidrule(r){4-6}  \cmidrule(r){7-11}  \cmidrule(r){12-14}
                              
& \multicolumn{1}{l}{}                                              & Method                 & \text{EL}$_3$ $\downarrow$        & MA $\downarrow$         & SS $\downarrow$     & Perplexity $\downarrow$  & \text{Rep}$_2$ $\downarrow$ & Div$_3$ $\uparrow$  & Coherence $\uparrow$ & MAUVE $\uparrow$ & Erasure & Generation & Avg. \\ \midrule
\multirow{9}{*}{\textbf{\rotatebox{90}{GPT-Neo-125M}}} & \multicolumn{1}{l}{}                                              & Original               & 0.212         & 0.789      & 0.587          & 27.69      & 0.123 & 0.923  & 0.566    & 0.702     & N/A     & N/A        & N/A  \\ 
\cline{2-14}
  & \multicolumn{1}{l}{\multirow{2}{*}{\textit{w/ MM}}} & TA        & 0.117      & 0.677    & 0.500          & \textbf{26.48}      & 0.346  & 0.737  & 0.554    & 0.276     &    4     &     4       &   4   \\
& \multicolumn{1}{l}{}                                              &  CD   & \underline{0.105}         & \underline{0.621}      & \textbf{0.419}          & 48.26      & 0.17   & 0.861  & \underline{0.570}      & 0.445      &   2      &      3      &   =2   \\
                              \cline{2-14}
  & \multirow{2}{*}{\textit{w/ RD}}                         & \cellcolor{red!40} GD    & 0              &  0           & 0.021               & 5.30           &  0.956      &   0.037     &   0.092        &    0.023        &     N/A    &  N/A   &  N/A    \\
&                                                                   & \cellcolor{red!40} KL          & 0           & 0.007      & 0.023          & 1.91      & 0.990  &  0.010  & 0.032     &  0.038     &   N/A  &  N/A          &   N/A   \\
                              \cline{2-14}
  & \multirow{3}{*}{\textit{w/o REF}}                        & \cellcolor{red!40} GA        & 0        & 0       & 0.012          & 2.36       & 0.990 & 0.010  & 0.051     &   0.011    &  N/A    & N/A   &   N/A   \\
&                                                                   & DI & 0.109         & 0.744       & 0.485          & 47.64      & \textbf{0.060}   & \textbf{0.965}  & 0.555     &  \underline{0.568}    &   3      &  2          &  =2    \\ \rowcolor{gray!40}  \cellcolor{white}
  &                    \cellcolor{white}                                               & EMSO (ours)                   & \textbf{0.065}         & \textbf{0.615}      & \underline{0.459}          & \underline{27.33}      & \underline{0.105}  & \underline{0.940}   & \textbf{0.572}      &    \textbf{0.610}   & 1        &  1   &  1    \\ \midrule  \midrule
\multirow{9}{*}{\textbf{\rotatebox{90}{GPT-Neo-1.3B}}} & \multicolumn{1}{l}{}                                              & Original               &    0.371           &  0.953           &      0.688          &       16.99     &   0.090     &   0.94     &        0.597   &      0.762      &   N/A      &    N/A         &   N/A    \\
\cline{2-14}
  & \multicolumn{1}{l}{\multirow{2}{*}{\textit{w/ MM}}} & TA        &    0.151     &    \underline{0.682}    &   0.467          &     \textbf{18.98}    &   0.377     &   0.733     &  \underline{0.552}         &     0.328       &    3     &     3       &    3  \\
  & \multicolumn{1}{l}{}                                              & CD   &   0.263           &     0.788        &      0.494          &     52.43       &   0.376     &    0.644    &     0.527      &     0.286       &   4      &      4      &   4   \\
                  \cline{2-14}
  & \multirow{2}{*}{\textit{w/ RD}}                         & \cellcolor{red!40} GD    &     0          &     0.002        &        0.006        &    535.82        &  0.038      &   0.922     &     0.011      &    0.02       &    N/A      &     N/A        &   N/A    \\
  &                                                                   & \cellcolor{red!40}KL          &     0         &      0       &     0          &   5.71     &  0.944  &   0.051     &     0.034      &     0.01       &   N/A      &    N/A         &    N/A   \\
                              \cline{2-14}
  & \multirow{3}{*}{\textit{w/o REF}}                        &\cellcolor{red!40} GA        &      0         &       0      &        0.061        & 3.49           &   0.984     &   0.023     &    0.066       &     0.015       &    N/A      &     N/A        &    N/A   \\
  &                                                                   & DI &       \underline{0.138}        &      0.751       &   \underline{0.457}             &       64.27    &    \textbf{0.057}    &    \textbf{0.967}    &      0.527     &    \underline{0.591}     &   2      &     2       & 2     \\ \rowcolor{gray!40} \cellcolor{white}
  &                 \cellcolor{white}                                                  & EMSO (ours)                   &   \textbf{0.135}   &   \textbf{0.623}    &   \textbf{0.431}         &    \underline{21.92}    &   \underline{0.090}     &  \underline{0.900}      &    \textbf{0.598}       &    \textbf{0.694}        &  1  & 1  & 1     \\ \midrule  \midrule
\multirow{9}{*}{\textbf{\rotatebox{90}{GPT-Neo-2.7B}}} & \multicolumn{1}{l}{}                                              & Original               &    0.377           &       0.966      &        0.744        &     13.83       &    0.083    &    0.953    &       0.599    &      0.790      &    N/A    &     N/A       &   N/A   \\
\cline{2-14}
  & \multirow{2}{*}{\textit{w/ MM}}                     & \cellcolor{red!40} TA        &    0.059           &     0.441        &       0.037         &     11.91       &   0.611     &   0.474     &    0.532        &   0.137        &   N/A     &     N/A       &   N/A   \\
  &                                                                   & CD   &    0.287           &     0.858        &      \underline{0.493}       &     36.12       &   0.348     &    0.686    &      0.476     &    0.371        &     3    &     3       &   3   \\
                              \cline{2-14}
  & \multirow{2}{*}{\textit{w/ RD}}                         & \cellcolor{red!40}GD    & 0              &      0      &     0.061  &       477.89     &    0.019    &   0.759     &      0.061     &      0.028      &   N/A      &     N/A       &   N/A   \\
&                                                                   &\cellcolor{red!40} KL          &       0        &      0       &     0.006     &     2.51       &    0.938    &   0.065     &    0.030   &    0.035        &    N/A     &     N/A       &    N/A  \\
                              \cline{2-14}
      & \multirow{3}{*}{\textit{w/o REF}}                        &\cellcolor{red!40} GA        &    0           &     0      &   0.033    &   2.07    & 0.992       &   0.012     &   0.032        &     0.010       &    N/A     &      N/A      &    N/A  \\
      &                                                                   & DI &      \textbf{0.225}         &       \underline{0.792}      &     0.525           &  \underline{22.78}          &   \textbf{0.059}     &   \textbf{0.952}    &    \underline{0.583}       &  \underline{0.692}          &    2     &   2     &  2    \\ \rowcolor{gray!40} \cellcolor{white}
  &           \cellcolor{white}                          & EMSO (ours)                   &     \underline{0.242}          &      \textbf{0.701}       &      \textbf{0.415}       &   \textbf{19.06}         &  \underline{0.095}      &   \underline{0.947}    &   \textbf{0.608}    &     \textbf{0.713}       &  1     &    1   &    1
\\  
\midrule
\bottomrule
\end{tabular}
}
\vspace{-0.2cm}
\caption{Experiment result of different TSM erasure methods on models with various scales. The best and the second-best results are highlighted in \textbf{bold} and \underline{underline} respectively. We rank the erasure and generation ability of an updated model by the times they achieve best/second best in corresponding metrics. We mark the collapse models with \colorbox{red!40}{red} and do not count collapse models in the ranking. }
\vspace{-0.5cm}
\label{tab:exp}

\end{table*}

\subsection{Experiment Setup}
We describe the models, data, baselines and evaluation metrics of our experiment in this section. 
Detailed experiment setup is in Appendix \ref{appd:details}.
\subsubsection{Test Model and Forget Set}
\textbf{Model.} We use the GPT-Neo model family (with 125M, 1.3B, 2.7B parameters) for evaluation because (\textit{i}) they are proven to memorize and emit training sample verbatim and (\textit{ii}) they are widely used in previous works \cite{jang2023knowledge, dong2024unmemorization, barbulescu2024each} to evaluate TSM.

\noindent\textbf{Data.} We use the dataset from Training Data Extraction Challenge as the forget set, which is a subset of Pile Corpora \cite{gao2020pile} and demonstrated to be easy-to-extract from pretrained GPT-Neo model family.
This dataset consists of 15,000 text sequences with a length of 200 tokens, which is ideal for evaluating TSM erasure with large forgetting requests. Compared with TOFU benchmark \cite{maini2024tofu}, the Extraction Challenge data is naturally memorized in the pretrainng stage of LLMs while TOFU simulates memorization by post hoc finetuning on fictional data.
Thus, we choose Extraction Challenge data instead of TOFU to test in a practical setting.

\vspace{-0.2cm}
\subsubsection{Comparison Methods}
We compare our method with seven state-of-the-art methods to reveal its effectiveness and model utility after model updating.
We divided the methods into three categories:
(\textit{i}) \emph{Updating with Memorized Model (w/ MM)}, which trains a model overfitting on forget set to act as a reference for forgetting, including \textbf{Task Arithmetic (TA)} \cite{ilharco2022editing} and \textbf{Contrastive Decoding (CD)} \cite{li2023contrastive}
(\textit{ii}) \emph{Updating with Retain Data (w/ RD)}, which assumes the existence of $D_r \in D \textbackslash D_f$ to maintain the model utility, including \textbf{Gradient Difference (GD)} \cite{liu2022continual} and \textbf{KL Divergence (KL)} \cite{wang2023kga}.
(\textit{iii}) \emph{Updating without Reference (w/o REF)}, which is a challenging setting that only requires the forget set and original model to complete the updation process, including \textbf{Gradient Ascent (GA)} \cite{jang2023knowledge} and \textbf{Deliberate Imagination (DI)} \cite{dong2024unmemorization}.
Our method lies in the w/o REF category.
The detailed description for comparison methods is in Appendix \ref{appd:desc}.

\subsubsection{Evaluation Metrics}

\fakeparagraph{Evaluation Metrics for Memorization.}
We use three different metrics to comprehensively evaluate the effectiveness of TSM erasure from exact memorization \cite{tirumala2022memorization} and approximate memorization \cite{ippolito2022preventing} perspectives:

(\textbf{\textit{i}}) Extraction Likelihood (EL) \cite{jang2023knowledge} compares n-grams overlap between generation from an updated model and the original continuation. The definition for EL is:
\begin{equation}
\begin{split}
    \text{EL}_{n}(\textbf{\textit{x}}) =& \frac{\sum_{i=1}^{p+q-n}\text{Overlap}_{n}(f_{\theta}(x_{1:i}), x_{i:p+q})   }{p+q-n},\\
    \text{Overlap}_n&(a,b) = \frac{|\text{n-gram}(a) \cap \text{n-gram}(b)|}{|\text{n-gram(a)}|}, \nonumber
\end{split}
\end{equation}
where $f_\theta(x_{1:i})$ is the generation from model $\theta$ given prefix $x_{1:i}$ and n-gram($\cdot$) is a list of n-grams for given sequence.

(\textbf{\textit{ii}}) Memorization Accuracy (MA) \cite{tirumala2022memorization} for quantifying the model memorization of given sequence $\textbf{\textit{x}}$:
\begin{align}
    \text{MA}(\textbf{\textit{x}}) = \frac{\sum_{i=p+1}^{p+q-1}\textbf{1}(\text{argmax}(P_{\theta,i})=x_i)}{q-1}. \nonumber
\end{align}

(\textbf{\textit{iii}}) Semantic Similarity (SS) for evaluating the semantic-level resemblance between model generation $f_\theta(x_{1:p})$ and original continuation $x_{p+1:q}$.
We extract semantic embedding from text sequence with MiniLM \cite{wang2020minilm} and compute the cosine similarity between the embeddings.
We introduce the metrics and datasets for model utility evaluation in Appendix \ref{appd:metrics}.

\begin{figure*}[t]
\centering
\includegraphics[width=\textwidth]{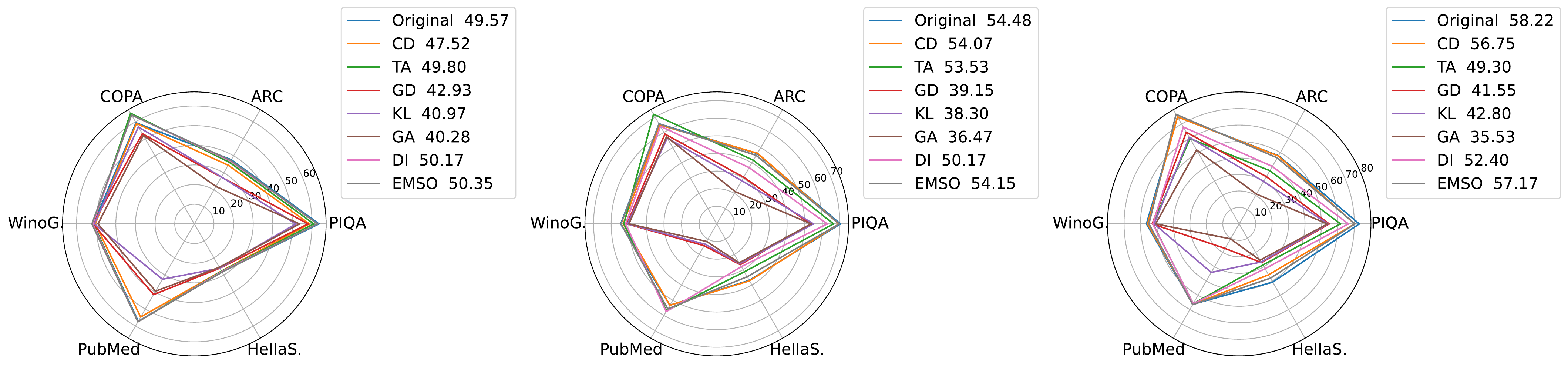}
\vspace{-0.5cm}
\caption{Experiment results for language reasoning ability evaluation with GPT-Neo-125M (left), 1.3B (middle), and 2.7B (right) as target model. We report the average accuracy of the updated model on all six tasks in the legend. Our EMSO achieves the best performance on all three models compared with baselines. }
\label{fig:NLI}
\vspace{-0.2cm}
\end{figure*}

\begin{figure}[h]
  \centering
  \includegraphics[width=0.9\linewidth]{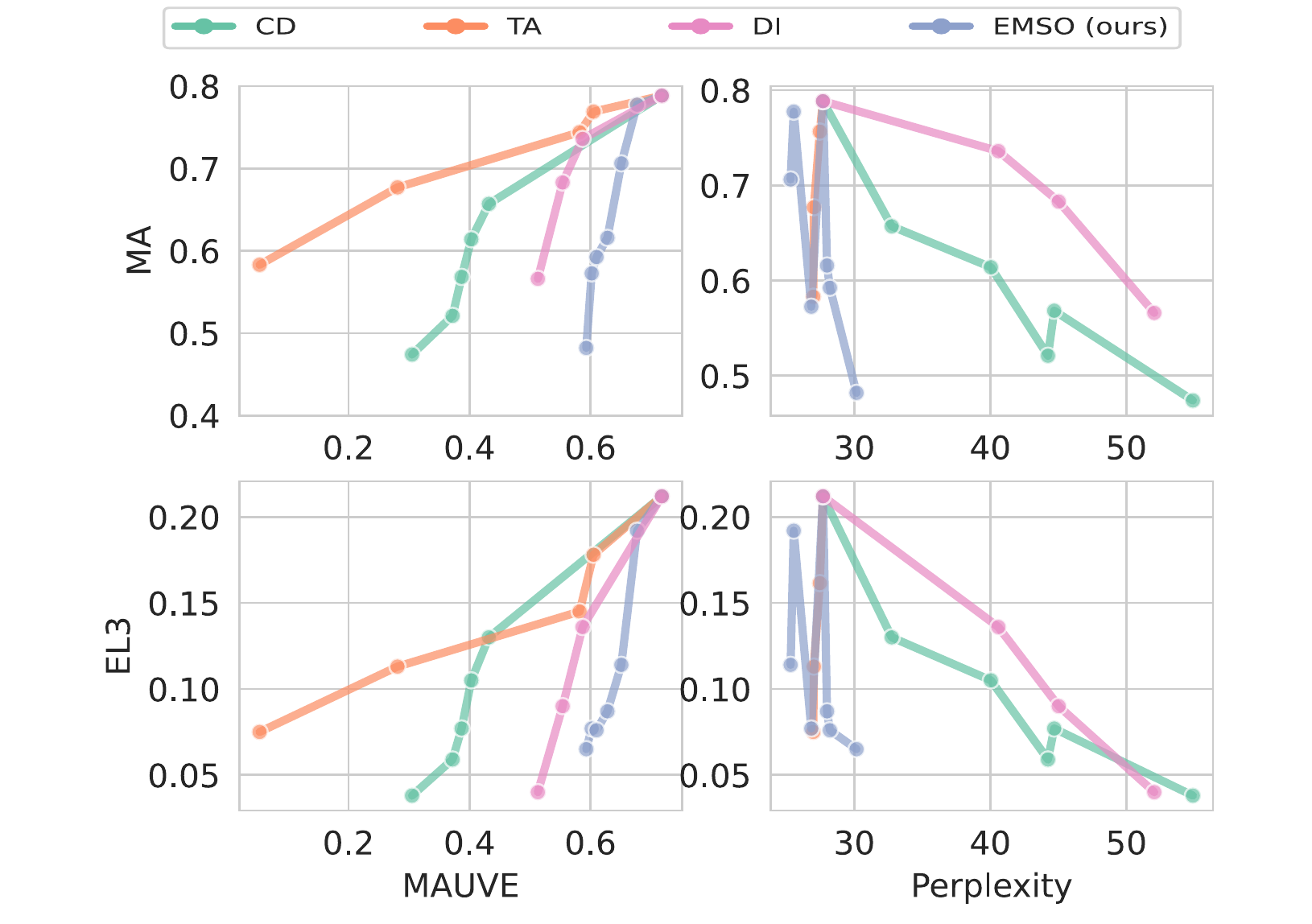}
  \vspace{-0.2cm}
  
  \caption{Illustration of erasure-utility trade-off for different methods on GPT-Neo-125M. We vary the erasure strength for different methods. The one with high MAUVE and low Perplexity while low on MA and $\text{EL}_3$ is considered better (\ie closer to the lower-right corner for MAUVE-MA and MAUVE-$\text{EL}_3$ figures and to the lower-left corner for Perplexity-MA and Perplexity-$\text{EL}_3$ figures). We do not plot the trade-off line for GD, KL, and GA here because they collapse before completing a single training epoch.}
  \label{fig:tradeoff} 
  \vspace{-0.6cm}
\end{figure}




\vspace{-0.3cm}
\subsection{Experiment Results}
\label{subsec:exp}
We show experiment results in Table \ref{tab:exp} together with the erasure-utility trade-off curve in Figure \ref{fig:tradeoff} and language reasoning ability evaluation in Figure \ref{fig:NLI}. 
We unveil the following five key insights: 
1) \textbf{Best performer.} Our method shows the best erasure-utility tradeoff among all competitors with erasure and utility all ranking first among 125M, 1.3B, and 2.7B models.
Moreover, Figure~\ref{fig:tradeoff} shows that our EMSO achieves comparable erasure effectiveness while sacrificing MAUVE by less than 0.1 and keeping Perplexity at the same level.
In contrast, CD, TA, and DI all compromise either MAUVE (TA, CD) or Perplexity (DI, CD) greatly to get satisfactory erasure performance.
2) \textbf{Model collapse.} All methods based on $\mathcal{L}_{NLL}$ \ie GD, KL, GA, completely collapse.
We categorize such collapse into two classes: text degeneration and gibberish generation.
Text degeneration means the model starts to repeat the same token, indicated by extremely low perplexity, high repetition and low MAUVE.
Gibberish generation means that the model outputs nonsense content, reflected by high perplexity and low MAUVE.
We observe that GD for the 1.3B and 2.7B model fall into gibberish generation while other $\mathcal{L}_{NLL}$-based methods show text degeneration.
The unsatisfied performance of $\mathcal{L}_{NLL}$-based methods demonstrates that optimizing $\mathcal{L}_{NLL}$ fails to keep the utility after erasure when processing massive  requests even if the retain data is available.
3) \textbf{Less affected reasoning ability.}
As shown in Figure \ref{fig:NLI}, compared with the significant deterioration in language generation ability after updation, the language reasoning ability of the updated model appears to be more stable.
Our EMSO still stands out among all competitors with average accuracy on six tasks dropping by 0.33\% and 1.05\% on 1.3B and 2.7B models and increasing by 0.78\% on 125M models.
The fluctuation of reasoning ability is within 3\% except for collapsed models.
It demonstrates that in LLMs, generation and reasoning ability are orthogonal to some extent and TSM erasure tends to destroy model capability in generation rather than reasoning. 
A similar phenomenon is also observed by \citet{barbulescu2024each}.
4) \textbf{TSM erasure is difficult to scale up.}
As model parameters scale from 125M to 2.7B, model memorization is stronger and harder to erase.
When we scale the original model from  125M to 2.7B, $\text{EL}_3$, MA and SS increase by 0.165, 0.177 and 0.157 respectively.
Moreover, with the same erasure strength, all erasure methods are less effective when applied to larger models.
For example, the 2.7B model achieves 0.792 in MA updated by DI, which is even higher than that in the smaller original 125M model.
5) \textbf{Evaluation bias.}
There exists a bias in different memorization metrics.
In the case of DI, it always performs better in $\text{EL}_3$ but weak in MA.
For instance, when updating the 2.7B model, DI excels our method by 0.017 in $\text{EL}_3$ but falls far behind in MA by 0.091.
Thus, it is necessary to use diverse metrics to evaluate erasure effectiveness comprehensively to avoid possible bias.
\vspace{-0.2cm}
\begin{table}[h]
\centering
\resizebox{\linewidth}{!}{
\begin{tabular}{c c c c c}
\toprule
\textbf{Method} & \textbf{\textbf{EL}$_3$}$\downarrow$ & \textbf{MA}$\downarrow$ & \textbf{Perplexity}$\downarrow$ & \textbf{MAUVE}$\uparrow$ \\
\midrule
\cellcolor{red!40} \textit{\textbf{Select \& NLL}} & 0.012& 0.008 & 5.218 & 0.006\\
\textit{\textbf{Random \& EM}} & 0.201& 0.785 & 29.695 & \textbf{0.701}\\
\textit{\textbf{w/o Dir}} & 0.080& \underline{0.590} & \underline{28.112} & 0.529\\
\textit{\textbf{Full \& EM}} & \underline{0.074}& 0.598 & 37.423 & 0.387\\
\textit{\textbf{Ours}} & \textbf{0.070}& \textbf{0.573} &  \textbf{26.831} &  \underline{0.602}\\
\bottomrule
\end{tabular}
}

\caption{Ablation study results using different variants. The best and the second-best result is highlighted in \textbf{bold} and \underline{underlined} text respectively. The collapsed model is marked with \colorbox{red!40}{red} and is not included when comparing results.} 
\vspace{-0.3cm}
\label{tab:ablation}
\end{table}

\subsection{Ablation Study}

To validate the necessity of every component in our proposed method, we conduct ablation studies with the following settings. \textbf{1) Select \& NLL} erases TSM of forget set data by updating top-2 salient blocks with $\mathcal{L}_{NLL}$. \textbf{2) Random \& EM} randomly selects blocks and finetunes them with $\mathcal{L}_{EM}$. \textbf{3) w/o Dir} selects top-k blocks with the largest $|\nabla \theta_{EM}|$ to update with $\mathcal{L}_{EM}$. \textbf{4) Full \& EM} updates the whole model with $\mathcal{L}_{EM}$.

The experiment results are reported in Table~\ref{tab:ablation}. 
Unsurprisingly, fine-tuning a model with $\mathcal{L}_{NLL}$ again leads to model collapse even if we only update the most salient weight.
Randomly picking weights brings little change to the model as all metrics stay close to the original model.
Moreover, our EMSO reduces MA by 0.017 and improves MAUVE by 0.073 compared with w/o Dir, demonstrating that taking the direction into consideration helps accurately locate blocks that are influential in updation and boost the erasure-utility trade-off. 
Fine-tuning the whole model with $\mathcal{L}_{EM}$ jeopardizes the model utility substantially to achieve similar erasure effectiveness of our method.
These results corroborate the function of each component of our proposed EMSO for improving erasure effectiveness while preserving model utility.

\begin{figure}[t]
  \centering
  \includegraphics[width=0.8\linewidth]{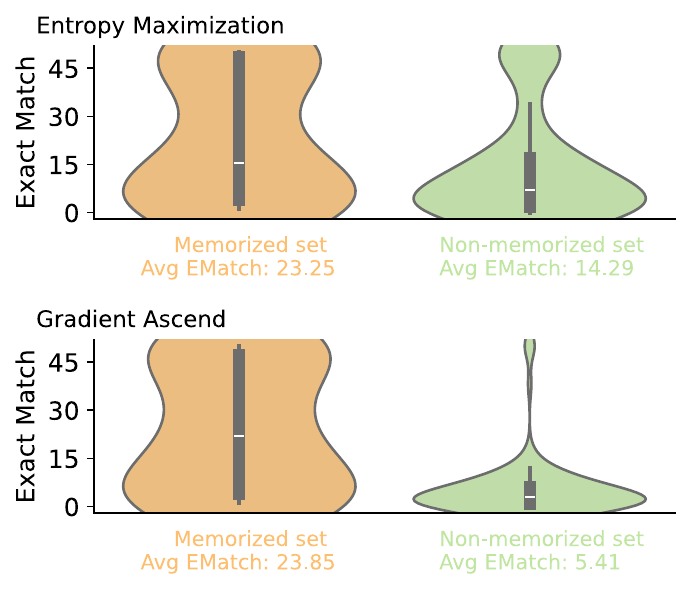}
  \vspace{-0.5cm}
  \caption{Illustration of the output change for memorized data and non-memorized data. We calculate the exact match between the original model and the updated model output given prefix from the forget set.}
  \label{fig:em} 
\vspace{-0.5cm}
\end{figure}

\vspace{-0.2cm}
\subsection{Discussion}
\label{discussion}
\vspace{-0.2cm}
\fakeparagraph{Analysis on Entropy Maximization Loss.}
We study different objectives' effectiveness on memorized and non-memorized data to better understand the reason why entropy maximization loss helps achieve a better trade-off between TSM erasure and model utility.
In practice, it is normal that the forget set consists of both memorized and non-memorized data because users are agnostic of whether their requests are memorized by a model in most cases.
As a simulation, we sample 20 memorized data and 100 non-memorized data from the mem-nonmem split of the Pile dataset\footnote{https://github.com/googleinterns/localizing-paragraph-memorization/tree/main/paragraphs/gpt-neo-125M/preds}.
This dataset quantifies model memorization with \textit{exact match} (EMatch) \cite{nasr2023scalable} which counts the number of matches between greedy-decoded tokens and ground truth tokens given the same prefix until the first mismatch.
Since the length of continuation in this dataset is 50, $\text{EMatch}=50$ is the maximum value and means the model repeats the text sequence verbatim.

We update model to forget the samples with $\gL_{EM}$ and $\gL_{NLL}$, respectively, and calculate the EMatch between outputs from the original model and the updated model given the prefix of requests.
As shown in Fig.~\ref{fig:em}, $\gL_{EM}$ and $\gL_{NLL}$ have similar effects on forgetting memorized data.
However, EMSO preserves non-memorized samples better while NLL-updated model changes completely on its generation on the non-memorized set.
We hypothesize that it is because NLL is a targeted objective for penalizing the probability of generating tokens while EMSO works in an untargeted fashion thus preserving the model's original ability on non-memorized data.

\fakeparagraph{Analysis on Selected Blocks.} We count the selection frequency of fine-tuning blocks and report the result in Table \ref{tab:freq}.
The selection process is conducted seven times before early stopping thus the max frequency should be seven. EMSO tends to select blocks at shallow layers with a total frequency of 11 out of 14, indicating that memorization is affected largely by shallow layers.
In addition, $\text{W}_v$ in shallow layers are most frequently selected, e.g., the $\text{L}_2 \text{W}_v\text{H}_{11}$\footnote{We name the blocks according to their position and function in the model with the format L\{layer number\}\{$\text{W}_k$, $\text{W}_q$, $\text{W}_v$, $\text{W}_o$, $\text{C}_{fc}$, $\text{C}_{proj}$\}H\{attention head number\}, where $\text{W}_k$, $\text{W}_q$, $\text{W}_v$, $\text{W}_o$ represent the linear transformation matrix for K,Q,V and output in attention mechanism and $\text{C}_{fc}$, $\text{C}_{proj}$ represent up projection and down projection matrix in MLP.} is selected in every round, suggesting the value matrix is most significant among the K, Q, V in the attention mechanism of LLMs regarding memorization.
This observation is consistent with \citet{he2024sparse}.
Moreover, we study the forward attention patterns of $\text{L}_2\text{H}_{11}$ to interpret its role in model memorization in Appendix \ref{appd: attention}.

\section{Conclusion}

This paper presents EMSO, a framework for textual sequence memorization erasure and better erasure-utility trade-off for LLMs when processing massive requests for verbatim memorization erasure.
We first show that entropy maximization is a better objective for TSM erasure than the commonly used NLL and label smoothing loss.
Then we provide theoretical analysis on the gradient of different updating objectives and find out entropy maximization provides more stable gradients, helps stabilize the updation process and avoids model collapse.
Moreover, we minimally invade the model by only updating blocks selected by a contrastive gradient metric for selecting salient blocks that optimize towards forgetting direction to get a better erasure-utility trade-off.
Our experiment results demonstrate the efficacy of our method compared with six baselines.
The discussion on erasure effectiveness for memorized and non-memorized data and the pattern of selected blocks also sheds light on studying TSM from data and model structure perspectives in LLMs.

\section*{Limitations}
In this work, we step forward to achieving a better erasure-utility trade-off when erasing model TSM about massive data.
However, several limitations still exist in our proposed method EMSO.
First, although EMSO performs best among all baseline methods in terms of TSM erasure, there is still a portion of requests that are not erased completely.
Second, despite the effectiveness of our contrastive gradient metric, it needs more memory to store both $\nabla\mathcal{L}_{EM}$ and $\nabla\mathcal{L}_{NLL}$ in the training stage, which limits its application on larger models.
Third, our methods greatly change the meaning of data in the forget set. 
Future work will focus on automatically detecting and editing only the privacy information in textual sequences while preserving the overall semantics of requests.

\section*{Ethics Statement}
The goal of our work is to protect user privacy from leaking by LLMs.
We redact the accurate privacy information used in the examples.
All the datasets used in this work are public.
We use the datasets consistent with their intended use.


\bibliography{acl_latex}

\begin{thebibliography}{59}
\providecommand{\natexlab}[1]{#1}

\bibitem[{Achiam et~al.(2023)Achiam, Adler, Agarwal, Ahmad, Akkaya, Aleman, Almeida, Altenschmidt, Altman, Anadkat et~al.}]{achiam2023gpt}
Josh Achiam, Steven Adler, Sandhini Agarwal, Lama Ahmad, Ilge Akkaya, Florencia~Leoni Aleman, Diogo Almeida, Janko Altenschmidt, Sam Altman, Shyamal Anadkat, et~al. 2023.
\newblock Gpt-4 technical report.
\newblock \emph{arXiv preprint arXiv:2303.08774}.

\bibitem[{Adebayo et~al.(2018)Adebayo, Gilmer, Muelly, Goodfellow, Hardt, and Kim}]{adebayo2018sanity}
Julius Adebayo, Justin Gilmer, Michael Muelly, Ian Goodfellow, Moritz Hardt, and Been Kim. 2018.
\newblock Sanity checks for saliency maps.
\newblock In \emph{Proceedings of the 32nd International Conference on Neural Information Processing Systems}, pages 9525--9536.

\bibitem[{Barbulescu and Triantafillou(2024)}]{barbulescu2024each}
George-Octavian Barbulescu and Peter Triantafillou. 2024.
\newblock To each (textual sequence) its own: Improving memorized-data unlearning in large language models.
\newblock \emph{arXiv preprint arXiv:2405.03097}.

\bibitem[{Bisk et~al.(2020)Bisk, Zellers, Gao, Choi et~al.}]{bisk2020piqa}
Yonatan Bisk, Rowan Zellers, Jianfeng Gao, Yejin Choi, et~al. 2020.
\newblock Piqa: Reasoning about physical commonsense in natural language.
\newblock In \emph{Proceedings of the AAAI conference on artificial intelligence}, volume~34, pages 7432--7439.

\bibitem[{Bourtoule et~al.(2021)Bourtoule, Chandrasekaran, Choquette-Choo, Jia, Travers, Zhang, Lie, and Papernot}]{bourtoule2021machine}
Lucas Bourtoule, Varun Chandrasekaran, Christopher~A Choquette-Choo, Hengrui Jia, Adelin Travers, Baiwu Zhang, David Lie, and Nicolas Papernot. 2021.
\newblock Machine unlearning.
\newblock In \emph{2021 IEEE Symposium on Security and Privacy (SP)}, pages 141--159. IEEE.

\bibitem[{Brown et~al.(2020)Brown, Mann, Ryder, Subbiah, Kaplan, Dhariwal, Neelakantan, Shyam, Sastry, Askell et~al.}]{brown2020language}
Tom Brown, Benjamin Mann, Nick Ryder, Melanie Subbiah, Jared~D Kaplan, Prafulla Dhariwal, Arvind Neelakantan, Pranav Shyam, Girish Sastry, Amanda Askell, et~al. 2020.
\newblock Language models are few-shot learners.
\newblock \emph{Advances in neural information processing systems}, 33:1877--1901.

\bibitem[{Carlini et~al.(2022)Carlini, Ippolito, Jagielski, Lee, Tramer, and Zhang}]{carlini2022quantifying}
Nicholas Carlini, Daphne Ippolito, Matthew Jagielski, Katherine Lee, Florian Tramer, and Chiyuan Zhang. 2022.
\newblock Quantifying memorization across neural language models.
\newblock In \emph{The Eleventh International Conference on Learning Representations}.

\bibitem[{Carlini et~al.(2021)Carlini, Tramer, Wallace, Jagielski, Herbert-Voss, Lee, Roberts, Brown, Song, Erlingsson et~al.}]{carlini2021extracting}
Nicholas Carlini, Florian Tramer, Eric Wallace, Matthew Jagielski, Ariel Herbert-Voss, Katherine Lee, Adam Roberts, Tom Brown, Dawn Song, Ulfar Erlingsson, et~al. 2021.
\newblock Extracting training data from large language models.
\newblock In \emph{30th USENIX Security Symposium (USENIX Security 21)}, pages 2633--2650.

\bibitem[{Chundawat et~al.(2023)Chundawat, Tarun, Mandal, and Kankanhalli}]{chundawat2023can}
Vikram~S Chundawat, Ayush~K Tarun, Murari Mandal, and Mohan Kankanhalli. 2023.
\newblock Can bad teaching induce forgetting? unlearning in deep networks using an incompetent teacher.
\newblock In \emph{Proceedings of the AAAI Conference on Artificial Intelligence}, volume~37, pages 7210--7217.

\bibitem[{Clark et~al.(2018)Clark, Cowhey, Etzioni, Khot, Sabharwal, Schoenick, and Tafjord}]{clark2018think}
Peter Clark, Isaac Cowhey, Oren Etzioni, Tushar Khot, Ashish Sabharwal, Carissa Schoenick, and Oyvind Tafjord. 2018.
\newblock Think you have solved question answering? try arc, the ai2 reasoning challenge.
\newblock \emph{arXiv preprint arXiv:1803.05457}.

\bibitem[{Dong et~al.(2024)Dong, Lin, Belkin, Huerta, and Vuli{\'c}}]{dong2024unmemorization}
Yijiang~River Dong, Hongzhou Lin, Mikhail Belkin, Ramon Huerta, and Ivan Vuli{\'c}. 2024.
\newblock Unmemorization in large language models via self-distillation and deliberate imagination.
\newblock \emph{arXiv preprint arXiv:2402.10052}.

\bibitem[{Eldan and Russinovich(2023)}]{eldan2023s}
Ronen Eldan and Mark Russinovich. 2023.
\newblock Who's harry potter? approximate unlearning in llms.
\newblock \emph{arXiv preprint arXiv:2310.02238}.

\bibitem[{Elhage et~al.(2021)Elhage, Nanda, Olsson, Henighan, Joseph, Mann, Askell, Bai, Chen, Conerly et~al.}]{elhage2021mathematical}
Nelson Elhage, Neel Nanda, Catherine Olsson, Tom Henighan, Nicholas Joseph, Ben Mann, Amanda Askell, Yuntao Bai, Anna Chen, Tom Conerly, et~al. 2021.
\newblock A mathematical framework for transformer circuits.
\newblock \emph{Transformer Circuits Thread}, 1:1.

\bibitem[{Fan et~al.(2024)Fan, Liu, Zhang, Wei, Wong, and Liu}]{fan2024salun}
Chongyu Fan, Jiancheng Liu, Yihua Zhang, Dennis Wei, Eric Wong, and Sijia Liu. 2024.
\newblock Salun: Empowering machine unlearning via gradient-based weight saliency in both image classification and generation.
\newblock In \emph{International Conference on Learning Representations}.

\bibitem[{Gao et~al.(2020)Gao, Biderman, Black, Golding, Hoppe, Foster, Phang, He, Thite, Nabeshima et~al.}]{gao2020pile}
Leo Gao, Stella Biderman, Sid Black, Laurence Golding, Travis Hoppe, Charles Foster, Jason Phang, Horace He, Anish Thite, Noa Nabeshima, et~al. 2020.
\newblock The pile: An 800gb dataset of diverse text for language modeling.
\newblock \emph{arXiv preprint arXiv:2101.00027}.

\bibitem[{Gao et~al.(2021)Gao, Yao, and Chen}]{gao2021simcse}
Tianyu Gao, Xingcheng Yao, and Danqi Chen. 2021.
\newblock {SimCSE}: Simple contrastive learning of sentence embeddings.
\newblock In \emph{Empirical Methods in Natural Language Processing (EMNLP)}.

\bibitem[{Graves et~al.(2021)Graves, Nagisetty, and Ganesh}]{graves2021amnesiac}
Laura Graves, Vineel Nagisetty, and Vijay Ganesh. 2021.
\newblock Amnesiac machine learning.
\newblock In \emph{Proceedings of the AAAI Conference on Artificial Intelligence}, volume~35, pages 11516--11524.

\bibitem[{Hamborg et~al.(2017)Hamborg, Meuschke, Breitinger, and Gipp}]{hamborg2017news}
Felix Hamborg, Norman Meuschke, Corinna Breitinger, and Bela Gipp. 2017.
\newblock news-please: a generic news crawler and extractor.

\bibitem[{He et~al.(2024)He, Li, Jiang, and Miller}]{he2024sparse}
Haoze He, Juncheng~Billy Li, Xuan Jiang, and Heather Miller. 2024.
\newblock Sparse matrix in large language model fine-tuning.
\newblock \emph{arXiv e-prints}, pages arXiv--2405.

\bibitem[{Huang et~al.(2022)Huang, Shao, and Chang}]{huang2022large}
Jie Huang, Hanyin Shao, and Kevin Chen-Chuan Chang. 2022.
\newblock Are large pre-trained language models leaking your personal information?
\newblock In \emph{Findings of the Association for Computational Linguistics: EMNLP 2022}, pages 2038--2047.

\bibitem[{Huang et~al.(2024)Huang, Yang, and Potts}]{huang2024demystifying}
Jing Huang, Diyi Yang, and Christopher Potts. 2024.
\newblock Demystifying verbatim memorization in large language models.
\newblock \emph{arXiv preprint arXiv:2407.17817}.

\bibitem[{Ilharco et~al.(2022)Ilharco, Ribeiro, Wortsman, Schmidt, Hajishirzi, and Farhadi}]{ilharco2022editing}
Gabriel Ilharco, Marco~Tulio Ribeiro, Mitchell Wortsman, Ludwig Schmidt, Hannaneh Hajishirzi, and Ali Farhadi. 2022.
\newblock Editing models with task arithmetic.
\newblock In \emph{The Eleventh International Conference on Learning Representations}.

\bibitem[{Ippolito et~al.(2022)Ippolito, Tram{\`e}r, Nasr, Zhang, Jagielski, Lee, Choquette-Choo, and Carlini}]{ippolito2022preventing}
Daphne Ippolito, Florian Tram{\`e}r, Milad Nasr, Chiyuan Zhang, Matthew Jagielski, Katherine Lee, Christopher~A Choquette-Choo, and Nicholas Carlini. 2022.
\newblock Preventing verbatim memorization in language models gives a false sense of privacy.
\newblock \emph{arXiv preprint arXiv:2210.17546}.

\bibitem[{Jagielski et~al.(2022)Jagielski, Thakkar, Tramer, Ippolito, Lee, Carlini, Wallace, Song, Thakurta, Papernot et~al.}]{jagielski2022measuring}
Matthew Jagielski, Om~Thakkar, Florian Tramer, Daphne Ippolito, Katherine Lee, Nicholas Carlini, Eric Wallace, Shuang Song, Abhradeep~Guha Thakurta, Nicolas Papernot, et~al. 2022.
\newblock Measuring forgetting of memorized training examples.
\newblock In \emph{The Eleventh International Conference on Learning Representations}.

\bibitem[{Jang et~al.(2023)Jang, Yoon, Yang, Cha, Lee, Logeswaran, and Seo}]{jang2023knowledge}
Joel Jang, Dongkeun Yoon, Sohee Yang, Sungmin Cha, Moontae Lee, Lajanugen Logeswaran, and Minjoon Seo. 2023.
\newblock Knowledge unlearning for mitigating privacy risks in language models.
\newblock In \emph{Proceedings of the 61st Annual Meeting of the Association for Computational Linguistics (Volume 1: Long Papers)}, pages 14389--14408.

\bibitem[{Jia et~al.(2023)Jia, Liu, Ram, Yao, Liu, Liu, Sharma, and Liu}]{jia2023model}
Jinghan Jia, Jiancheng Liu, Parikshit Ram, Yuguang Yao, Gaowen Liu, Yang Liu, Pranay Sharma, and Sijia Liu. 2023.
\newblock Model sparsity can simplify machine unlearning.
\newblock In \emph{Proceedings of the 37th International Conference on Neural Information Processing Systems}, pages 51584--51605.

\bibitem[{Jin et~al.(2019)Jin, Dhingra, Liu, Cohen, and Lu}]{jin2019pubmedqa}
Qiao Jin, Bhuwan Dhingra, Zhengping Liu, William Cohen, and Xinghua Lu. 2019.
\newblock Pubmedqa: A dataset for biomedical research question answering.
\newblock In \emph{Proceedings of the 2019 Conference on Empirical Methods in Natural Language Processing and the 9th International Joint Conference on Natural Language Processing (EMNLP-IJCNLP)}, pages 2567--2577.

\bibitem[{Kaplan et~al.(2020)Kaplan, McCandlish, Henighan, Brown, Chess, Child, Gray, Radford, Wu, and Amodei}]{kaplan2020scaling}
Jared Kaplan, Sam McCandlish, Tom Henighan, Tom~B Brown, Benjamin Chess, Rewon Child, Scott Gray, Alec Radford, Jeffrey Wu, and Dario Amodei. 2020.
\newblock Scaling laws for neural language models.
\newblock \emph{arXiv preprint arXiv:2001.08361}.

\bibitem[{Kassem et~al.(2023)Kassem, Mahmoud, and Saad}]{kassem2023preserving}
Aly Kassem, Omar Mahmoud, and Sherif Saad. 2023.
\newblock Preserving privacy through dememorization: An unlearning technique for mitigating memorization risks in language models.
\newblock In \emph{Proceedings of the 2023 Conference on Empirical Methods in Natural Language Processing}, pages 4360--4379.

\bibitem[{Li et~al.(2024)Li, Pan, Gopal, Yue, Berrios, Gatti, Li, Dombrowski, Goel, Mukobi et~al.}]{liwmdp}
Nathaniel Li, Alexander Pan, Anjali Gopal, Summer Yue, Daniel Berrios, Alice Gatti, Justin~D Li, Ann-Kathrin Dombrowski, Shashwat Goel, Gabriel Mukobi, et~al. 2024.
\newblock The wmdp benchmark: Measuring and reducing malicious use with unlearning.
\newblock In \emph{Forty-first International Conference on Machine Learning}.

\bibitem[{Li et~al.(2023)Li, Holtzman, Fried, Liang, Eisner, Hashimoto, Zettlemoyer, and Lewis}]{li2023contrastive}
Xiang~Lisa Li, Ari Holtzman, Daniel Fried, Percy Liang, Jason Eisner, Tatsunori Hashimoto, Luke Zettlemoyer, and Mike Lewis. 2023.
\newblock Contrastive decoding: Open-ended text generation as optimization.
\newblock In \emph{The 61st Annual Meeting Of The Association For Computational Linguistics}.

\bibitem[{Liu et~al.(2022)Liu, Liu, and Stone}]{liu2022continual}
Bo~Liu, Qiang Liu, and Peter Stone. 2022.
\newblock Continual learning and private unlearning.
\newblock In \emph{Conference on Lifelong Learning Agents}, pages 243--254. PMLR.

\bibitem[{Liu et~al.(2024)Liu, Dou, Tan, Tian, and Jiang}]{liu2024towards}
Zheyuan Liu, Guangyao Dou, Zhaoxuan Tan, Yijun Tian, and Meng Jiang. 2024.
\newblock Towards safer large language models through machine unlearning.
\newblock \emph{arXiv preprint arXiv:2402.10058}.

\bibitem[{Loshchilov and Hutter(2018)}]{loshchilov2018decoupled}
Ilya Loshchilov and Frank Hutter. 2018.
\newblock Decoupled weight decay regularization.
\newblock In \emph{International Conference on Learning Representations}.

\bibitem[{Maini et~al.(2024)Maini, Feng, Schwarzschild, Lipton, and Kolter}]{maini2024tofu}
Pratyush Maini, Zhili Feng, Avi Schwarzschild, Zachary~C Lipton, and J~Zico Kolter. 2024.
\newblock Tofu: A task of fictitious unlearning for llms.
\newblock \emph{arXiv preprint arXiv:2401.06121}.

\bibitem[{Mantelero(2013)}]{mantelero2013eu}
Alessandro Mantelero. 2013.
\newblock The eu proposal for a general data protection regulation and the roots of the ‘right to be forgotten’.
\newblock \emph{Computer Law \& Security Review}, 29(3):229--235.

\bibitem[{Meng et~al.()Meng, Sharma, Andonian, Belinkov, and Bau}]{mengmass}
Kevin Meng, Arnab~Sen Sharma, Alex~J Andonian, Yonatan Belinkov, and David Bau.
\newblock Mass-editing memory in a transformer.
\newblock In \emph{The Eleventh International Conference on Learning Representations}.

\bibitem[{Merity et~al.(2016)Merity, Xiong, Bradbury, and Socher}]{merity2016pointer}
Stephen Merity, Caiming Xiong, James Bradbury, and Richard Socher. 2016.
\newblock Pointer sentinel mixture models.
\newblock In \emph{International Conference on Learning Representations}.

\bibitem[{M{\"u}ller et~al.(2019)M{\"u}ller, Kornblith, and Hinton}]{muller2019does}
Rafael M{\"u}ller, Simon Kornblith, and Geoffrey~E Hinton. 2019.
\newblock When does label smoothing help?
\newblock \emph{Advances in neural information processing systems}, 32.

\bibitem[{Nasr et~al.(2023)Nasr, Carlini, Hayase, Jagielski, Cooper, Ippolito, Choquette-Choo, Wallace, Tram{\`e}r, and Lee}]{nasr2023scalable}
Milad Nasr, Nicholas Carlini, Jonathan Hayase, Matthew Jagielski, A~Feder Cooper, Daphne Ippolito, Christopher~A Choquette-Choo, Eric Wallace, Florian Tram{\`e}r, and Katherine Lee. 2023.
\newblock Scalable extraction of training data from (production) language models.
\newblock \emph{arXiv preprint arXiv:2311.17035}.

\bibitem[{Rafailov et~al.(2024)Rafailov, Sharma, Mitchell, Manning, Ermon, and Finn}]{rafailov2024direct}
Rafael Rafailov, Archit Sharma, Eric Mitchell, Christopher~D Manning, Stefano Ermon, and Chelsea Finn. 2024.
\newblock Direct preference optimization: Your language model is secretly a reward model.
\newblock \emph{Advances in Neural Information Processing Systems}, 36.

\bibitem[{Roemmele et~al.(2011)Roemmele, Bejan, and Gordon}]{roemmele2011choice}
Melissa Roemmele, Cosmin~Adrian Bejan, and Andrew~S Gordon. 2011.
\newblock Choice of plausible alternatives: An evaluation of commonsense causal reasoning.
\newblock In \emph{2011 AAAI Spring Symposium Series}.

\bibitem[{Sakaguchi et~al.(2021)Sakaguchi, Bras, Bhagavatula, and Choi}]{sakaguchi2021winogrande}
Keisuke Sakaguchi, Ronan~Le Bras, Chandra Bhagavatula, and Yejin Choi. 2021.
\newblock Winogrande: An adversarial winograd schema challenge at scale.
\newblock \emph{Communications of the ACM}, 64(9):99--106.

\bibitem[{Shi et~al.(2024)Shi, Lee, Huang, Malladi, Zhao, Holtzman, Liu, Zettlemoyer, Smith, and Zhang}]{shi2024muse}
Weijia Shi, Jaechan Lee, Yangsibo Huang, Sadhika Malladi, Jieyu Zhao, Ari Holtzman, Daogao Liu, Luke Zettlemoyer, Noah~A Smith, and Chiyuan Zhang. 2024.
\newblock Muse: Machine unlearning six-way evaluation for language models.
\newblock \emph{arXiv preprint arXiv:2407.06460}.

\bibitem[{Su et~al.(2022)Su, Lan, Wang, Yogatama, Kong, and Collier}]{su2022contrastive}
Yixuan Su, Tian Lan, Yan Wang, Dani Yogatama, Lingpeng Kong, and Nigel Collier. 2022.
\newblock A contrastive framework for neural text generation.
\newblock \emph{Advances in Neural Information Processing Systems}, 35:21548--21561.

\bibitem[{Tirumala et~al.(2022)Tirumala, Markosyan, Zettlemoyer, and Aghajanyan}]{tirumala2022memorization}
Kushal Tirumala, Aram Markosyan, Luke Zettlemoyer, and Armen Aghajanyan. 2022.
\newblock Memorization without overfitting: Analyzing the training dynamics of large language models.
\newblock \emph{Advances in Neural Information Processing Systems}, 35:38274--38290.

\bibitem[{Touvron et~al.(2023)Touvron, Martin, Stone, Albert, Almahairi, Babaei, Bashlykov, Batra, Bhargava, Bhosale et~al.}]{touvron2023llama}
Hugo Touvron, Louis Martin, Kevin Stone, Peter Albert, Amjad Almahairi, Yasmine Babaei, Nikolay Bashlykov, Soumya Batra, Prajjwal Bhargava, Shruti Bhosale, et~al. 2023.
\newblock Llama 2: Open foundation and fine-tuned chat models.
\newblock \emph{arXiv preprint arXiv:2307.09288}.

\bibitem[{Vaswani et~al.(2017)Vaswani, Shazeer, Parmar, Uszkoreit, Jones, Gomez, Kaiser, and Polosukhin}]{vaswani2017attention}
Ashish Vaswani, Noam Shazeer, Niki Parmar, Jakob Uszkoreit, Llion Jones, Aidan~N Gomez, {\L}ukasz Kaiser, and Illia Polosukhin. 2017.
\newblock Attention is all you need.
\newblock \emph{Advances in neural information processing systems}, 30.

\bibitem[{Wang(2021)}]{mesh-transformer-jax}
Ben Wang. 2021.
\newblock {Mesh-Transformer-JAX: Model-Parallel Implementation of Transformer Language Model with JAX}.
\newblock \url{https://github.com/kingoflolz/mesh-transformer-jax}.

\bibitem[{Wang et~al.(2023)Wang, Chen, Yuan, Zeng, Wong, and Yin}]{wang2023kga}
Lingzhi Wang, Tong Chen, Wei Yuan, Xingshan Zeng, Kam-Fai Wong, and Hongzhi Yin. 2023.
\newblock Kga: A general machine unlearning framework based on knowledge gap alignment.
\newblock In \emph{Proceedings of the 61st Annual Meeting of the Association for Computational Linguistics (Volume 1: Long Papers)}, pages 13264--13276.

\bibitem[{Wang et~al.(2024{\natexlab{a}})Wang, Zeng, Guo, Wong, and Gottlob}]{wang2024selective}
Lingzhi Wang, Xingshan Zeng, Jinsong Guo, Kam-Fai Wong, and Georg Gottlob. 2024{\natexlab{a}}.
\newblock Selective forgetting: Advancing machine unlearning techniques and evaluation in language models.
\newblock \emph{arXiv preprint arXiv:2402.05813}.

\bibitem[{Wang et~al.(2020)Wang, Wei, Dong, Bao, Yang, and Zhou}]{wang2020minilm}
Wenhui Wang, Furu Wei, Li~Dong, Hangbo Bao, Nan Yang, and Ming Zhou. 2020.
\newblock Minilm: Deep self-attention distillation for task-agnostic compression of pre-trained transformers.
\newblock \emph{Advances in Neural Information Processing Systems}, 33:5776--5788.

\bibitem[{Wang et~al.(2024{\natexlab{b}})Wang, Wu, He, Chen, and McAuley}]{wang2024large}
Yu~Wang, Ruihan Wu, Zexue He, Xiusi Chen, and Julian McAuley. 2024{\natexlab{b}}.
\newblock Large scale knowledge washing.
\newblock \emph{arXiv preprint arXiv:2405.16720}.

\bibitem[{Wu et~al.()Wu, Li, Xu, Dong, Wu, Bian, and Xiong}]{wu2023depn}
Xinwei Wu, Junzhuo Li, Minghui Xu, Weilong Dong, Shuangzhi Wu, Chao Bian, and Deyi Xiong.
\newblock Depn: Detecting and editing privacy neurons in pretrained language models.
\newblock In \emph{The 2023 Conference on Empirical Methods in Natural Language Processing}.

\bibitem[{Yao et~al.(2023)Yao, Xu, and Liu}]{yao2023large}
Yuanshun Yao, Xiaojun Xu, and Yang Liu. 2023.
\newblock Large language model unlearning.
\newblock \emph{arXiv preprint arXiv:2310.10683}.

\bibitem[{Yona and Greenfeld(2021)}]{yona2021revisiting}
Gal Yona and Daniel Greenfeld. 2021.
\newblock Revisiting sanity checks for saliency maps.
\newblock In \emph{eXplainable AI approaches for debugging and diagnosis.}

\bibitem[{Zellers et~al.(2019)Zellers, Holtzman, Bisk, Farhadi, and Choi}]{zellers2019hellaswag}
Rowan Zellers, Ari Holtzman, Yonatan Bisk, Ali Farhadi, and Yejin Choi. 2019.
\newblock Hellaswag: Can a machine really finish your sentence?
\newblock In \emph{Proceedings of the 57th Annual Meeting of the Association for Computational Linguistics}, pages 4791--4800.

\bibitem[{Zhang et~al.(2023)Zhang, Chen, Liu, and He}]{zhang2023composing}
Jinghan Zhang, Shiqi Chen, Junteng Liu, and Junxian He. 2023.
\newblock Composing parameter-efficient modules with arithmetic operation.
\newblock In \emph{Thirty-seventh Conference on Neural Information Processing Systems}.

\bibitem[{Zhang et~al.(2024)Zhang, Lin, Bai, and Mei}]{zhang2024negative}
Ruiqi Zhang, Licong Lin, Yu~Bai, and Song Mei. 2024.
\newblock Negative preference optimization: From catastrophic collapse to effective unlearning.
\newblock \emph{arXiv preprint arXiv:2404.05868}.

\end{thebibliography}
\clearpage
\appendix

\section{Difference in Minimizer for Label Smoothing Loss and Entropy Maximization Loss}
\label{appd:mini}
\begin{figure}[h]
  \centering
  \includegraphics[width=\linewidth]{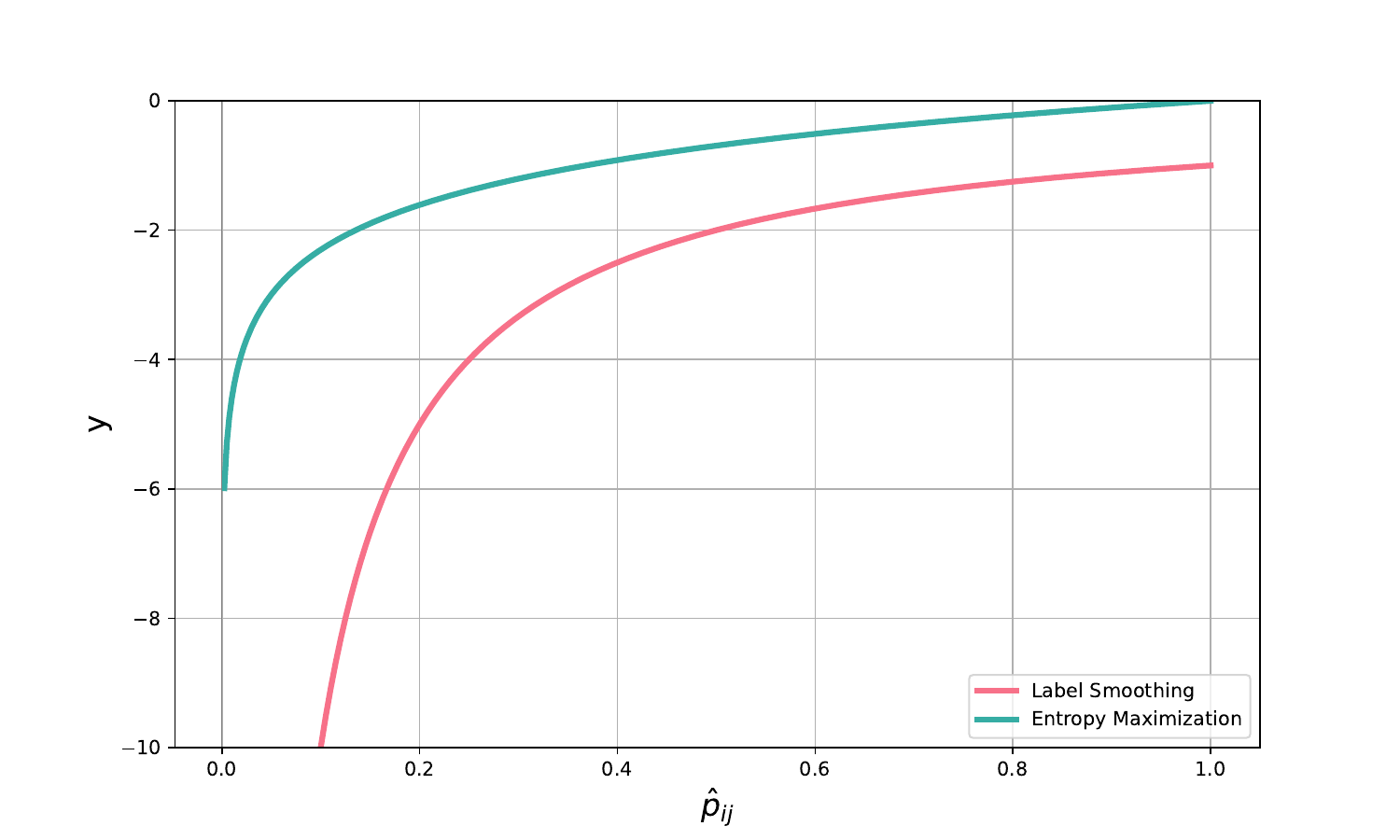}
  \caption{Illustration of minimizer scale and gradient difference between label smoothing loss and entropy maximization loss. The scale and gradient of entropy maximization loss are much smaller than those of label smoothing loss, indicating entropy maximization provides a more stable updating process.}
  \label{fig:mini} 
  \vspace{-15pt}
\end{figure}

Following the notion in Section \ref{subsec:em}. The gradient ascent loss aims to maximize the following objective:
\begin{align}
    \gL_{NLL}=-\log \hat p_{ij},
\end{align}
As $\hat p_{ij}$ is the output of the softmax function with $h_{ij}$ as input, we take the derivative of the loss function with respect to the input $h_{ij}$.
\begin{align}
    \frac{\partial \gL_{NLL}}{\partial h_{ik}}&=- \frac{1}{\hat p_{ij}}\frac{\partial \hat p_{ij}}{\partial h_{ik}}
    \label{eq:nll}
\end{align}
Comparing Equation \ref{eq:nll} and Equation \ref{equ:ls_grad}, they share the same minimizer $- \frac{1}{\hat p_{ij}}$, indicating gradient acsend loss also have greater gradient which deteriorates stable optimization.

\section{Influence of the number of selected blocks}
\label{appd:k}
In this section, we study how the number of selected blocks would affect the erasure-utility trade-off.
The experiment results are shown in Table \ref{tab:block}.
\begin{table}[h]
\centering
\resizebox{\linewidth}{!}{
\begin{tabular}{c c c c c}
\toprule
\textbf{Num. Blocks} & \textbf{$\textbf{EL}_3$} & \textbf{MA} & \textbf{Perplexity} & \textbf{MAUVE} \\
\midrule
\textbf{1} & 0.131& 0.720 & 26.86 & 0.694\\
\textbf{2} & \textbf{0.065}& \textbf{0.615} & 27.33 & \textbf{0.701}\\
\textbf{3} & 0.078& 0.641 & \textbf{25.01} & 0.657\\
\textbf{4} & 0.094& 0.666 & 26.38 & 0.612\\

\bottomrule
\end{tabular}
}
\caption{Experiment Results on different numbers of selected blocks}
\label{tab:block}
\end{table}
Fine-tuning 2 blocks with entropy maximization objective leads to the best erasure-utility trade-off with the lowest $\text{EL}_3$ and MA, and highest MAUVE.
Only updating one block leads to more information leakage as $\text{EL}_3$ and MA increase by 0.066 and 0.105, respectively.
Interestingly, selecting more updating blocks not only does not help erase TSM but also impairs model utility indicated by dropping on MAUVE.


\section{Inplementation Details}
\label{appd:details}
We report all hyperparameter settings and hardware information in our experiments.
For updating GPT-Neo-125M, 1.3B, and 2.7B models, we set the batch size to 64, 16, and 8, respectively, and set gradient accumulation to 1, 4, 8 for simulating the same updation steps.
To make a fair comparison, we set the learning rate as $1e^{-5}$ for all methods with AdamW as optimizer \cite{loshchilov2018decoupled}.
We set early stop criteria for the updating process to be perplexity increased by 3\% on the WikiText-103 validation set.
We require the updating process to complete at least one epoch to make sure all forgetting requests are processed.
We use 10,000 textual sequences randomly sampled from CC News \cite{hamborg2017news} as retain data for GD and KL.
For w/ MM methods, we train the memorized model for 10 epochs on the forget set.
For methods that require erasure strength $\gamma$ setting, \ie TA, CD, DI, we set $\gamma$ to 0.05, 0.3, 3, respectively for results in Table \ref{tab:exp}.
For the experiment result in Figure \ref{fig:tradeoff}, we take $\gamma$ from range $[0.04,0.05,0.08,0.1]$, $[0.5,0.6,0.7,0.8]$, $[3,5,8,10]$ for TA, CD, DI, respectively and report EMSO results after epochs from 1 to 7 since training epoch is the key parameter for controlling erasure strength in EMSO.
We conduct all our experiments on a single NVIDIA Tesla A100 80GB GPU.

\begin{figure*}[t]
\centering
\includegraphics[width=\textwidth]{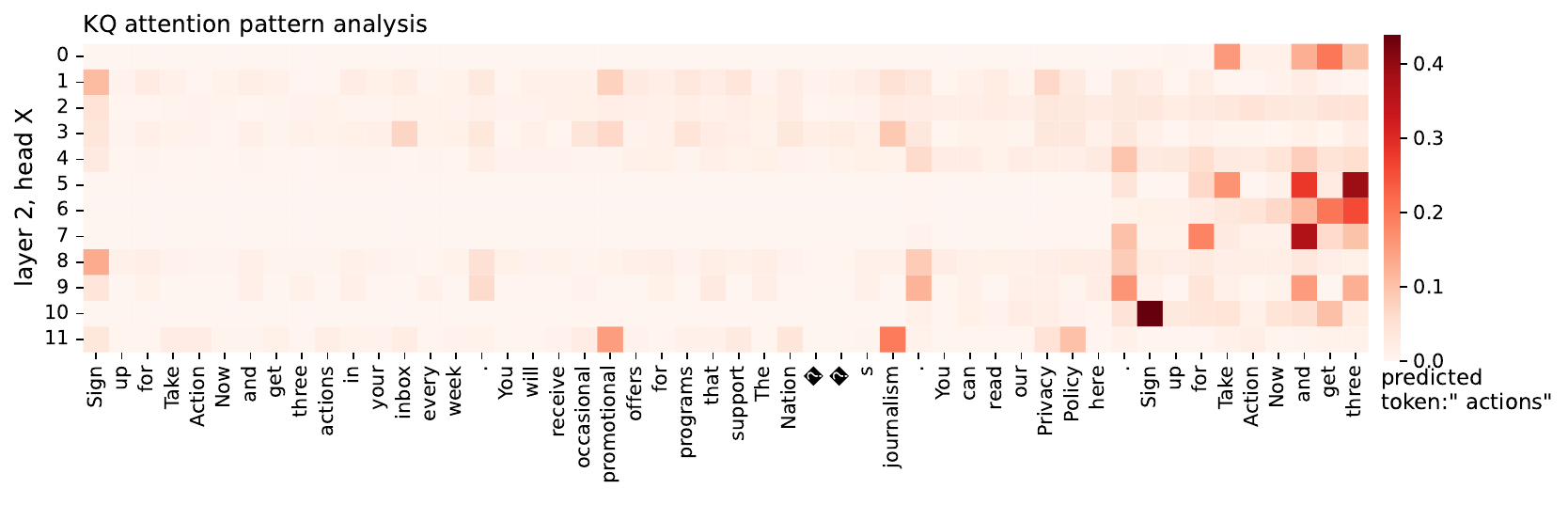}

\caption{KQ attention pattern analysis across all attention head in layer 2. We especially pay attention to head 11 because it is the most frequently selected head during weight selection process in EMSO. }
\label{fig:kq}
\vspace{-12pt}
\end{figure*}
\section{Detailed Description of Comparison Methods}
\label{appd:desc}
We introduce the details about comparison methods which can be categorized as \emph{\textbf{w/ MM}}, \emph{\textbf{w/ RD}}, and \emph{\textbf{w/o REF}}:

\noindent \emph{1. The \textbf{w/ MM} methods are as follows:}

\fakeparagraph{Task Arithmetic (TA)} \cite{ilharco2022editing}, which erases TSM by subtracting the memorization model weight from the original model and the process can be formularized as:
\begin{equation}
    \theta_{\text{TA}} = \theta_o - \gamma \cdot \theta_{\text{Memo}}.
\end{equation}
where $\theta_{\text{TA}}$, $\theta_o$, $\theta_{\text{Memo}}$ are the parameters of the updated model, original model, and memorized model, respectively. 
$\gamma$ controls the erasure strength, larger $\gamma$ indicates model memorization about the forget set is removed more completely at the cost of more severe utility destruction.

\fakeparagraph{Contrastive Decoding (CD)} \cite{li2023contrastive}, which steers original model output away from the memorized model with the following operation:
\begin{equation}
    P_\theta(x_i|x<i) = \text{softmax}(z_t-\gamma \cdot \text{RELU}(z_t^{\text{memo}}-z_t)),
\end{equation}
where $P_\theta(x_i|x<i)$ is the next token prediction probability distribution of the updated model.
$z_t$, $z_t^\text{memo}$ are the logits from the original model and the memorized model.
$\text{RELU}$ is the activation function. $\gamma$ controls erasure strength.

\noindent \emph{2. The \textbf{w/ RD} methods are as follows:}

\fakeparagraph{Gradient Difference (GD)} \cite{liu2022continual}, which increases NLL on forget set while decreasing it on retain set.

\fakeparagraph{KL Divergence (KL)} \cite{wang2023kga},
which preserves model utility by restraining KL-divergence between the output distribution of the updated model and the original model on the retain set.

\noindent \emph{3. The \textbf{w/o REF} methods are as follows:}

\fakeparagraph{Gradient Ascent (GA)} \cite{jang2023knowledge},
which penalizes each label token from text sequences in forget set.


\fakeparagraph{Deliberate Imagination (DI)} \cite{dong2024unmemorization},
which uses a label-smoothing loss to increase the sampling possibility on all tokens in the vocabulary other than memorized ones and can be formulated as:
\begin{equation}
    L = \sum_{t=1}^{T}\mathcal{L}_{CE}(z_t + \gamma\textbf{\text{1}}_{i,t}, z_s), 
\end{equation}
where $\mathcal{L}_{CE}$ is cross entropy, $\textbf{\text{1}}_{i,t}$ is all-ones vector except for ground truth ones, and $z_t, z_s$ are logits from teacher model and student model.
$\gamma$ is erasure strength.

\section{Utility Evaluation Metrics and Datasets}
\label{appd:metrics}
We test the language generation ability and reasoning ability of the updated model since they are the two most important functions of LLMs.

(\textit{i}) For language generation ability evaluation, we randomly sample 5,000 text sequences from Wikitext-103 dataset \cite{merity2016pointer} and take the first 32 tokens as input to language model for open generation.
Following \citet{su2022contrastive}, we use perplexity, diversity, repetition, MAUVE, Semantic Coherence for evaluating the generation quality.
We calculate perplexity with GPT-J-6B \cite{mesh-transformer-jax} and calculate semantic coherence with SimCSE \cite{gao2021simcse}.

(\textit{ii}) For language reasoning ability evaluation, we use a suite of popular NLP reasoning tasks, namely Piqa \cite{bisk2020piqa}, ARC-Easy \cite{clark2018think}, COPA \cite{roemmele2011choice}, PubmedQA \cite{jin2019pubmedqa}, Winogrande \cite{sakaguchi2021winogrande} and Hellaswag \cite{zellers2019hellaswag} for comprehensive evaluation.

\section{KQ Pattern Analysis on Most Frequently Selected Blocks}
\label{appd: attention}

\begin{table}[h]
\centering
\scalebox{0.8}{
\begin{tabularx}{\columnwidth}{@{\hspace{1cm}}c@{\hspace{2.5cm}}c}
\bottomrule
  Block Name & Frequency \\
  \hline      
$\text{L}_2\text{W}_v\text{H}_{11}$ & 7 \\

$\text{L}_{11}\text{C}_{proj}$ & 3  \\

$\text{L}_3\text{W}_o\text{H}_{2}$ & 2 \\

$\text{L}_3\text{W}_v\text{H}_{11}$ & 1 \\

$\text{L}_1\text{W}_o\text{H}_{8}$ & 1 \\ 
\bottomrule
& 
\end{tabularx}
}
\vspace{-0.5cm}
\caption{The frequency of updating blocks selection. }
\label{tab:freq}
\end{table}

As shown in Figure \ref{fig:kq}, the value matrix of layer 2, attention head 11 is selected in every weight selection round.
To better understand the mechanism of how this particular block affects memorization, we conduct analysis on its attention pattern at the inference stage.
To be specific, we study which previous tokens the attention head 11 in layer 2 pays attention to when decoding at the current step by calculating the normalized inner product of "keys" $k$ and queries $q$ in forward pass activations of attention block when provided certain memorized samples.
As shown in Figure \ref{fig:kq}, $\text{L}_2\text{H}_{11}$ pays the most attention to "promotional" and "journalism" in the given prefix.
Compared with other tokens such as "sign" and "you", $\text{L}_2\text{H}_{11}$ apparently concentrates on rare tokens with complex semantics in the input text sequence at the inference stage, indicating rare tokens might be functional in LLM memorization.

\end{document}